\documentclass{article}

\usepackage{PRIMEarxiv}

\usepackage[utf8]{inputenc}
\usepackage[T1]{fontenc}
\usepackage{amsmath,amssymb,amsfonts}
\usepackage{booktabs}
\usepackage{tabularx}
\usepackage{multirow}
\usepackage{enumitem}
\usepackage{xcolor}
\usepackage{hyperref}
\usepackage[numbers,compress]{natbib}
\usepackage{longtable}
\usepackage{array}
\usepackage{makecell}

\hypersetup{
    colorlinks=true,
    linkcolor=blue,
    citecolor=blue,
    urlcolor=blue
}

\newcommand{\E}{\mathbb{E}}
\newcommand{\Prob}{\mathbb{P}}
\newcommand{\D}{\mathcal{D}}
\newcommand{\T}{\mathcal{T}}

\newcommand{\PiSet}{\Pi}
\newcommand{\A}{\mathcal{A}}

\newcommand{\Env}{\mathcal{E}}
\newcommand{\Model}{\widehat{\mathcal{E}}}
\newcommand{\state}{s}
\newcommand{\obs}{o}
\newcommand{\act}{a}
\newcommand{\hist}{h}
\newcommand{\traj}{\tau}
\newcommand{\J}{J}
\newcommand{\Jhat}{\widehat{J}}
\newcommand{\rank}{\operatorname{rank}}

\newcommand{\ECE}{\operatorname{ECE}}

\newcolumntype{Y}{>{\raggedright\arraybackslash}X}
\title{How Should World Models Be Evaluated for Embodied Decision-Making?\\
A Decision-Making-Centric Position}

\author{
  Yang Yu$^{1,2,}$\thanks{Corresponding Author: yuy@nju.edu.cn}\ \ ,  Shiyuan Zhang$^{1,2}$, Yifei Sheng$^{1,2}$, Haoxiang Ren$^{1,2}$, Haoxin Lin$^{1,2,3}$ \\
  $^1$ National Key Laboratory for Novel Software Technology, Nanjing University, Nanjing, China \\
  $^2$ School of Artificial Intelligence, Nanjing University, Nanjing, China \\
  $^3$ Cirquar Technologies, Nanjing, China
}

\begin{document}

\maketitle

\begin{abstract}
World models have become a central abstraction in modern AI. The term now refers to several different objects: action-conditioned environment models, latent imagination models, future-video predictors, interactive neural simulators, latent predictive representations, and synthetic-data engines. Evaluation has broadened along with the term. Recent papers measure video realism, perceptual similarity, instruction following, physical plausibility, policy ranking, executability, planning success, and downstream policy improvement. This produces both metric diversity and a recurring problem of \emph{claim/evidence mismatch}: papers sometimes make a stronger claim about what their model is useful for than their evaluation can establish.

This paper surveys the recent literature and argues that, for models presented as world models for embodied decision-making, the more decisive issue is not whether the model generates visually convincing videos, but whether it supports reliable interventional reasoning, policy evaluation, planning, and policy optimization under intervention, policy-induced distribution shift, and long-horizon rollout. This principle is actually not new: it is the \emph{objective-mismatch} and \emph{decision-aware model learning} lesson from model-based reinforcement learning, which already established that predictive accuracy is often a poor proxy for control utility. This paper presents (i) a survey of \emph{evaluation practice} in the recent generative-world-model literature, (ii) a diagnosis, read off that survey, that this literature has re-encountered objective mismatch without adopting the tools developed to address it, and (iii) an operational evaluation framework and benchmark protocol. We organize the survey using an L0--L7 ladder spanning visual plausibility to policy optimization utility, noting that the levels cut across several orthogonal axes and so form an evidential hierarchy rather than a single scalar. The framework foregrounds interventional action fidelity, closed-loop rollout validity, reward/value prediction, policy-ranking agreement, optimization lift, model exploitability, and uncertainty calibration, with a minimal feasible reporting set for real-robot settings.
\end{abstract}

\section{Introduction}

World models have become an active theme in contemporary AI. In one line of work, inherited from model-based reinforcement learning, a world model is a learned dynamics model used for planning, imagination, policy evaluation, or policy optimization \citep{ha2018world,janner2019mbpo,hafner2023dreamerv3,wu2023daydreamer}. In another line, recent embodied video-generation models are described as world models because they generate plausible future observations from text, images, videos, or actions \citep{yang2024unisim,eva2024,worldsimbench2024,worldmodelbench2025,nvidia2025pbench}. A third line studies latent predictive representations, where the model predicts future embeddings rather than pixels \citep{assran2025vjepa2,vjepa21_2026,leworldmodel2026}. A fourth line uses generative models as synthetic-data engines or executable video planners for robot learning \citep{dreamgen2025,drema2025,robomaster2026,robowmbench2026}.

This proliferation has been productive, but it has also made evaluation ambiguous. Some papers evaluate pixel reconstruction or distributional video quality using MSE, PSNR, SSIM, LPIPS, FID, and FVD. Others evaluate instruction following or physical plausibility using VLM judges, physical QA, or human preference. Others use final policy success after training inside the model. Still others use the correlation between world-model-estimated policy success and real or simulator success \citep{worldgym2025,tseng2025scalable,dworldeval2026,worldarena2026}. These evaluations are not interchangeable. A model can be a strong video generator while being a poor environment model for control; conversely, a latent predictive model may be useful for planning without ever producing photorealistic pixels.

This observation is not new in spirit. In model-based reinforcement learning, the \emph{objective-mismatch} problem already documented that one-step predictive likelihood is frequently uncorrelated with downstream control performance \citep{lambert2020objmismatch}, and the \emph{decision-aware model learning} line, including value-aware model learning and the value-equivalence principle, argued that models should be evaluated and trained against their eventual decision use rather than against reconstruction \citep{modelgradient,farahmand2017vaml,grimm2020value,grimm2021proper,wei2024unified}. We therefore aim at carrying this underlying principle into the modern generative-world-model setting, where the dominant evaluation culture has drifted back toward reconstruction- and perception-style metrics, and specifically to (i) survey what the recent literature actually measures, (ii) show, from that survey, that the field has re-encountered objective mismatch without using the existing decision-aware vocabulary or diagnostics, and (iii) provide an operational framework and protocol that make the missing evidence concrete and reportable.

Our thesis is therefore conditional rather than universal. We do \emph{not} claim that every system called a world model should be judged by policy optimization. If the intended use is future-video generation, then video quality and semantic plausibility are legitimate primary targets. The difficulty arises when evidence appropriate for one claim is used to support a stronger one. If a model is presented as a world model for embodied decision-making, then the question we find most informative is:
\begin{quote}
\emph{What would happen, in task-relevant terms, if the agent took these actions from this history?}
\end{quote}

A skeptical reader might object that visual quality, semantic plausibility, or human preference can correlate with downstream utility in some settings, or that some systems called world models are simply not intended for control. We agree on both points. Our claim is therefore comparative rather than eliminative. We do not deny the usefulness of lower-level metrics or of purely generative world-model research. We argue that, for models whose stated aim is embodied decision-making, action-, outcome-, and policy-level evaluations usually provide stronger evidence than artifact quality alone.

This view is reflected in environment-model work on counterfactual learning, policy-conditioned models, full-horizon rollout, and generalizable embodied decision-making \citep{chen2023acem,chen2024pcm,lin2026admv2,zhang2024whale}, and in recent benchmarks that explicitly separate perceptual quality from functional utility \citep{worldgym2025,dworldeval2026,worldarena2026,robowmbench2026}. (Throughout, where cited titles use ``counterfactual'', for example ACEM's ``counterfactual environment-model learning'', we read it as the interventional notion we adopt in Section~2.6, not as the abduction-based, fixed-noise sense.)

This paper focuses on \emph{world models claimed for embodied decision-making}: policy evaluation, planning, policy optimization, safety testing, and related uses. It is therefore neither anti-video-metric nor anti-VLM-judge. Our claim is that, for this use case, video and semantic metrics are often better interpreted as lower-level or auxiliary diagnostics unless the claimed use is itself purely generative.
We make four contributions in this paper:
\begin{enumerate}[leftmargin=1.5em]
    \item We provide a paper-by-paper survey of the recent world-model literature, organized by \emph{what each paper actually evaluates}, and connect it explicitly to the older objective-mismatch and decision-aware model learning literature.
    \item Reading off that survey, we quantify and document a recurring failure mode: \emph{claim/evidence mismatch}, in which lower-level evidence is informally taken to support stronger decision-making claims, while interventional and exploitability diagnostics are largely absent.
    \item We organize the literature using an L0--L7 world-model evaluation ladder, ranging from visual plausibility to policy optimization utility, while clarifying that the levels span several orthogonal axes and therefore constitute an evidential hierarchy rather than a single scalar scale.
    \item We propose a decision-making-centric evaluation framework and benchmark protocol built around interventional branches, policy-induced distribution shift, full-horizon outcome fidelity, policy-ranking agreement, optimization lift, exploitability, and uncertainty calibration, including a minimal feasible variant for real-robot settings.
\end{enumerate}

\section{Background and Symbols}

\subsection{A brief genealogy of the term}

In the setting most relevant to this paper, the closest ancestor of the phrase ``world model'' is the \emph{environment model} tradition in model-based control and reinforcement learning. In that tradition, the core object is an action-conditioned predictive model of dynamics, rewards, and sometimes uncertainty, used to answer questions of the form: if the agent takes action \(\act\) from state or history \(\hist\), what is likely to happen next, and what consequences will this have for return or task completion? In this narrower sense, the conceptual target is already interventional and decision-theoretic: the model is valuable because it supports planning, policy evaluation, or policy improvement \citep{ha2018world,janner2019mbpo,hafner2023dreamerv3,wu2023daydreamer}.

The label ``world model'' became visible through work that paired compact representation learning with latent dynamics and a controller \citep{ha2018world}. In that formulation, the world model did not mean a photorealistic simulator of everything in the environment. It referred to an internal predictive model of environment evolution, usually in a latent state space, sufficient to support control. Dreamer-style methods and real-robot extensions such as DayDreamer preserved this interpretation: the world model is primarily a tool for imagined rollouts, value estimation, and policy learning \citep{hafner2023dreamerv3,wu2023daydreamer}.

The term broadened as embodied AI and large-scale generative modeling matured. A first shift came from \emph{future-observation prediction}. In many embodied and web-scale settings, video is the most accessible supervisory signal, while explicit action or reward labels may be scarce or heterogeneous. Models that predict future frames, videos, or multimodal continuations began to be described as world models, especially when they were used to forecast how scenes evolve under language, image, video, or action conditions \citep{eva2024,worldsimbench2024,worldmodelbench2025,nvidia2025pbench}. In this broader usage, the word ``world'' often refers to the model's ability to generate plausible futures, even if the model is not directly evaluated as a policy-evaluation or policy-optimization tool.

A second shift came from \emph{interactive neural simulators}. Once action-conditioned video models became capable of autoregressive rollout, it became natural to reuse them as surrogate environments. Systems such as UniSim, Vid2World, IRASim, WorldGym, and WorldArena sit in this intermediate region: they are still generative models of future observations, but they are also queried as if they were interactive environments \citep{yang2024unisim,vid2world2026,irasim2025,worldgym2025,worldarena2026}. This blurs the boundary between ``video predictor'' and ``world simulator,'' and it is one reason evaluation becomes ambiguous: the same model can be evaluated visually in one paragraph and functionally in the next.

A third shift came from \emph{latent predictive representations}. In JEPA-style and related approaches, the modeling target is not pixel reconstruction but future latent structure. These methods argue, implicitly or explicitly, that a useful world model may be one that predicts abstract, planning-relevant representations rather than photorealistic images \citep{assran2025vjepa2,vjepa21_2026,leworldmodel2026}. This line breaks the assumption that the natural output of a world model is a video, and it sharpens the distinction between \emph{observational fidelity} and \emph{decision-relevant sufficiency}.

A fourth shift came from \emph{world models as synthetic-data engines or executable planners}. Here the model is often not used as a general-purpose simulator in the classical sense. Instead, it may generate robot videos that are converted into actions, augment datasets with imagined trajectories, or produce demonstrations that improve downstream learning \citep{dreamgen2025,drema2025,robomaster2026,robowmbench2026}. The phrase ``world model'' still refers to a model of environmental evolution, but the operational role is neither pure dynamics learning nor pure video generation. It is instrumental: generate useful training signal, executable plans, or interventional data.

Viewed this way, the current literature contains not one but several partially overlapping world-model traditions. The same term now covers at least six research objects: action-conditioned environment models, latent imagination models, future-video predictors, interactive neural simulators, latent predictive representations, and synthetic-data engines. These objects overlap, but they are not identical, and their evaluations need not be identical either.

\begin{table*}[h]
\centering
\small
\begin{tabularx}{\textwidth}{p{1.3cm}p{3cm}p{4cm}p{3.4cm}X}
\toprule
\textbf{Phase} & \textbf{Object commonly called a world model} & \textbf{Why this usage emerged} & \textbf{Representative works} & \textbf{Typical evaluation emphasis} \\
\midrule
I & Action-conditioned environment model & Planning, control, off-policy evaluation, imagination-based learning & \citep{ha2018world,janner2019mbpo,hafner2023dreamerv3,wu2023daydreamer} & Policy return, sample efficiency, model-based planning, value estimation \\
\midrule
II & Latent imagination model & Need for compact long-horizon predictive state under partial observability & \citep{ha2018world,hafner2023dreamerv3,wu2023daydreamer} & Latent rollout quality, return prediction, downstream control \\
\midrule
III & Future-video predictor & Availability of large-scale video data; embodied tasks naturally expressed as future visual prediction & \citep{eva2024,worldsimbench2024,worldmodelbench2025,nvidia2025pbench} & Video fidelity, semantics, physical plausibility \\
\midrule
IV & Interactive neural simulator & Autoregressive action-conditioned video models reused as surrogate environments & \citep{yang2024unisim,vid2world2026,irasim2025,worldgym2025,worldarena2026} & Closed-loop rollout quality, policy ranking, planning success \\
\midrule
V & Latent predictive representation & Reaction against pixel-centric evaluation; emphasis on abstraction and planning relevance & \citep{assran2025vjepa2,vjepa21_2026,leworldmodel2026} & Planning, probing, transfer, dense correspondence, value-relevant features \\
\midrule
VI & Synthetic-data engine / executable planner & Generative models used instrumentally to create trajectories, demonstrations, or plans & \citep{dreamgen2025,drema2025,robomaster2026,robowmbench2026,eva_model2026} & Downstream policy lift, executability, action recovery, imitation gains \\
\bottomrule
\end{tabularx}
\caption{A brief genealogy of the term ``world model.'' The same label now covers several partially overlapping objects, which helps explain why evaluation practices have diverged.}
\label{tab:genealogy}
\end{table*}

\subsection{Objective mismatch and decision-aware model learning}

Our position is best understood as a continuation of an existing debate in model-based reinforcement learning, rather than as a new claim. We make this lineage explicit, because the recent generative-world-model literature has largely re-derived the question without connecting to the prior answers.

\paragraph{Objective mismatch.}
The objective-mismatch problem observes that the objective used to \emph{train} a dynamics model, typically one-step predictive likelihood or reconstruction error, is different from, and often uncorrelated with, the objective that actually matters, namely downstream control performance \citep{lambert2020objmismatch}. A globally accurate model is sufficient for planning but is neither necessary nor, under finite capacity, achievable; conversely, a model that is accurate only where it matters for the task can support strong control. The modern reliance on FID, FVD, PSNR, or VLM physics scores as primary world-model metrics is, in our reading, a re-instance of exactly this mismatch.

\paragraph{Decision-aware and value-equivalent model learning.}
A complementary line argues that model learning should be made aware of its decision role. Value-aware model learning weights model errors by their effect on value estimates rather than treating all prediction errors equally \citep{farahmand2017vaml}. The value-equivalence principle formalizes a related intuition: two models are equivalent for a set of policies and value functions if they induce the same Bellman updates, so a model need only be accurate in the value-relevant subspace, which generally shrinks as the policy/function set grows \citep{grimm2020value,grimm2021proper}. We use this principle in its precise form where a reward/value set is defined, and only as an informal analogy in settings (such as decoder-free latent models) where no such set is specified. A recent survey unifies these threads under the heading of decision-aware MBRL \citep{wei2024unified}. The practical consequence for evaluation is direct: the right notion of model fidelity is relative to a declared set of policies, value functions, and outcomes, which is precisely the ``decision contract'' we propose in Section~5.

\paragraph{Model exploitation.}
A third established phenomenon is that planners and optimizers actively seek out regions where the model is overoptimistic, so that improving average accuracy does not prevent catastrophic failures at the optimizer's chosen points \citep{janner2019mbpo,lambert2020objmismatch}. Our later exploitability gap (\(\mathrm{XGap}\), Section~5) is a measurement of exactly this effect, which is not a new concept, only as an evaluation quantity that the recent generative-world-model literature rarely reports (Section~3.6 quantifies this).

Against this background, the present paper does not propose a new training principle. It argues that the \emph{evaluation} practices of the recent generative-world-model literature have drifted away from these lessons, and it offers a concrete framework and protocol for realigning them.

\subsection{A narrow and a broad reading of ``world model''}

The survey suggests it is useful to distinguish two readings of the term.

\paragraph{Narrow reading (decision-theoretic).}
A world model is an action-conditioned predictive model that supports interventional reasoning about trajectories, outcomes, and values. Here the primary questions are interventional: what would happen if the agent chose this action sequence, and how good would that be?

\paragraph{Broad reading (predictive or generative).}
A world model is any model that predicts or generates future states of the world, whether in pixels, video, latent space, symbolic form, or another representation. Here the model may or may not be directly used for planning or control.

Neither reading is illegitimate, but they place pressure on different evaluations. The narrow reading foregrounds policy relevance. The broad reading licenses a wider range of predictive or generative metrics. Much of the confusion in the literature arises when evidence collected under the broad reading is discussed as if it also settled questions posed by the narrow one.

The broadening of the term can be understood as the result of several converging pressures. First, \emph{data availability} shifted: large, varied video corpora are easier to obtain than large corpora of reset-matched action branches with reliable action and reward annotation, which encourages observation-centric training and evaluation. Second, \emph{modeling scale} shifted: video diffusion and transformer-based generative models became strong enough that predicting future observations began to look like a plausible route to modeling the world, especially in embodied settings where the world is primarily encountered through images and video \citep{eva2024,worldsimbench2024}. Third, \emph{use cases diversified}: some researchers want a surrogate evaluator, some a data engine, some a planner, some a representation learner, and some an open-ended video generator, and the same family of models can serve more than one role but not always equally well. Fourth, \emph{interfaces blurred}: a model that predicts future observations under actions can be treated as a simulator; one that predicts latent futures can be treated as a planning substrate; one that generates useful rollouts can be treated as a data engine. Once these uses became common, the older environment-model language and the newer generative-world-model language merged.

\subsection{Controlled-process view}

Let \(\Env\) denote the real environment, which may be an MDP, a POMDP, a real robot task family, or a trusted simulator used as ground truth. A trajectory is
\[
    \traj = (\obs_0,\act_0,r_0,\obs_1,\act_1,r_1,\ldots,\obs_H).
\]
A history at time \(t\) is
\[
    \hist_t=(\obs_0,\act_0,\ldots,\obs_t).
\]
A policy \(\pi\) maps histories to actions. Its discounted return is
\[
    \J_{\Env}(\pi)
    =
    \E_{\traj \sim \Env,\pi}
    \left[
        \sum_{t=0}^{H-1}\gamma^t r_t
    \right].
\]

We use \(\Model\) as a deliberately broad abstraction for the learned object under evaluation:
\[
    \widehat{\traj}_{t:t+H}
    \sim
    \Model(\cdot \mid \hist_t, \act_{t:t+H-1}, c),
\]
where \(c\) may include language instructions, goals, embodiment information, camera viewpoint, latent state, or policy context. Depending on the paper, the output may be pixels, latent states, rewards, values, success probabilities, uncertainty estimates, action proposals, or some combination of these.

This notation is more general than the classical transition model \(P(\state_{t+1},r_t\mid \state_t,\act_t)\). Some surveyed models output only videos; some output only latent states; some output only scores or judgments. The notation abstracts over these cases so that differences in evaluation can be discussed in a common language. The key point is therefore not output modality. The key question is what decisions the model is meant to support, and what evidence has been provided for that use.

\subsection{Decision uses and evidential burdens}

We use \(U\) to denote the intended \emph{decision use} of the model:
\[
    U \in
    \{
        \text{prediction},
        \text{policy evaluation},
        \text{planning},
        \text{policy optimization},
        \text{synthetic data},
        \text{representation}
    \}.
\]
Different uses place pressure on different properties.

\begin{itemize}[leftmargin=1.5em]
    \item If \(U=\text{prediction}\), visual fidelity, semantics, and physical plausibility may be primary.
    \item If \(U=\text{policy evaluation}\), policy ranking and value calibration become central.
    \item If \(U=\text{planning}\), action sensitivity and long-horizon outcome fidelity become important.
    \item If \(U=\text{policy optimization}\), exploitability, distribution shift, and uncertainty become important.
    \item If \(U=\text{synthetic data}\), the main question is whether generated rollouts improve downstream learning.
    \item If \(U=\text{representation}\), the emphasis shifts toward abstraction, probing, and planning relevance rather than decoding quality.
\end{itemize}

This is why a single use-independent world-model score is difficult to defend. Different tasks and uses make different demands, a point that the value-equivalence literature makes precise: model fidelity is only well defined relative to a set of policies and value functions \citep{grimm2020value,grimm2021proper}.

\subsection{Open-loop, closed-loop, intervention, and identifiability}

Several distinctions are central for the rest of the paper.

\paragraph{Open-loop vs.\ closed-loop.}
In open-loop evaluation, the model predicts future observations from a ground-truth history. In closed-loop evaluation, the policy acts on model-generated histories and the model must remain coherent under its own predictions. Many world models look strong open-loop and degrade sharply closed-loop.

\paragraph{Observational vs.\ interventional prediction.}
A model can predict likely futures under the behavior-policy data distribution without correctly modeling the consequences of \emph{alternative} actions. For decision-making, the relevant object is
\[
    \Prob_{\Env}(\obs_{t+1:t+H}\mid \hist_t, do(\act_{t:t+H-1})),
\]
not merely
\[
    \Prob_{\Env}(\obs_{t+1:t+H}\mid \hist_t).
\]
We use \(do(\cdot)\) in the interventional sense (Pearl's second rung): we ask about the distribution of futures when actions are \emph{set} externally, holding the history fixed. We deliberately avoid the stronger word ``counterfactual'' in its technical sense, which would additionally require abduction over a fixed realization of exogenous noise; the evaluations we propose are interventional, and where the term ``counterfactual'' appears (including in cited titles) we mean the interventional, action-branching notion.

\paragraph{Identifiability, and its connection to pipeline entanglement.}
Estimating \(\Prob_{\Env}(\obs_{t+1:t+H}\mid \hist_t, do(\act_{t:t+H-1}))\) from logged data is not automatic. It generally requires (i) no unobserved confounding between the data-collection policy and the dynamics, so that conditioning on the recorded history is sufficient to block back-door paths, and (ii) adequate action coverage, so that the relevant action branches are supported in the data or can be actively queried. In fully instrumented simulators and many tabletop robot setups where the logged history captures the controller's information state and actions are recorded without hidden side-channels, the interventional and observational conditionals coincide and the quantity is identified by conditioning. This favorable case is, however, less common than the framing of the recent literature suggests. In the increasingly popular regime of learning world models from \emph{human} video (e.g., generalist video world models and executable-video pipelines), the ``action'' is not recorded but \emph{inferred} by an inverse-dynamics or retargeting model, and the recorded history is not the executed controller's information state. There the interventional target is generally \emph{not} identified even with environment resets, because the video-to-action map is itself a learned, error-prone component. This is the same phenomenon we elsewhere call pipeline entanglement: when executability or action recovery depends on inverse dynamics, retargeting, or success checkers \citep{robowmbench2026,robomaster2026,eva_model2026}, an interventional claim is a claim about the whole pipeline, not about the world model in isolation. In partially observed or open-world settings these assumptions can likewise fail, and interventional error must then be measured through active intervention (resets and executed action branches) rather than inferred from passive logs. Our benchmark protocol (Section~6) is designed so that the strongest interventional claims are made only where such interventions are actually performed and the action interface is directly controlled rather than inferred.

\paragraph{Outcome fidelity.}
For models intended to inform policy choice or policy improvement, preserving task-relevant outcomes (e.g., rewards, success predicates, progress variables, and constraints) is often more important than preserving every pixel. This is, informally, the value-equivalence intuition restated at the level of evaluation \citep{farahmand2017vaml,grimm2020value}.

\paragraph{Representation sufficiency.}
A model may fail to reconstruct visual detail yet still preserve the abstract state structure needed for planning. Conversely, a model may reconstruct visually plausible futures while failing to encode the causal distinctions that matter for action selection \citep{vafa2024implicit}.

\section{What Is Being Evaluated?}

This section is intentionally inventory-like. Tables~\ref{tab:bench-survey}--\ref{tab:latent-survey} enumerate recent works, grouped by what they most directly evaluate (the levels L0--L7 are described in the next section). The goal is solely to separate distinct evaluation cultures that are often collapsed under the single phrase ``world-model evaluation.''

\paragraph{Level-assignment rule.}
To keep the tables auditable, we use a single mechanical rule: a paper is listed at a level if and only if it reports at least one \emph{quantitative} metric whose target matches that level's core question (Table~\ref{tab:ladder}), regardless of how much emphasis the paper places on it. We do not weight by prominence and do not exercise a separate ``dominant evidence'' judgment; the level set in the last column is therefore intended to be reconstructable from each paper's reported metrics. Purely qualitative demonstrations (e.g., ``we show controllability in Figure~X'' without a number) are noted in the text but not counted as occupying a level.

\subsection{Benchmark and diagnostic literature}

\scriptsize
\begin{longtable}{p{2.55cm}p{2.45cm}p{4.25cm}p{4.55cm}p{1.35cm}}
\caption{Benchmark and diagnostic literature. The last column lists the main levels of the L0--L7 ladder occupied by the benchmark.}\label{tab:bench-survey}\\
\toprule
\textbf{Work} & \textbf{Claimed object} & \textbf{What is actually evaluated} & \textbf{Representative metrics or outputs} & \textbf{Main levels} \\
\midrule
\endfirsthead
\toprule
\textbf{Work} & \textbf{Claimed object} & \textbf{What is actually evaluated} & \textbf{Representative metrics or outputs} & \textbf{Main levels} \\
\midrule
\endhead
\textbf{EVA-Bench} \citep{eva2024}
& embodied future-video anticipation
& Offline embodied video prediction from current visual context and language; action description, next-step prediction, how-to generation, finish-thinking, and future-video quality
& BLEU, METEOR, ROUGE-L, CIDEr, SPICE, CLIPScore, GPT-4o score; SC, BC, MS, FVD, GCE; EVA-Score
& L1--L3 \\
\midrule
\textbf{WorldSimBench} \citep{worldsimbench2024}
& video generation models as world simulators
& Explicit perceptual evaluation in open embodied, driving, and manipulation settings; implicit closed-loop evaluation in Minecraft, CARLA, and CALVIN
& Human Preference Evaluator scores; route completion, infractions, driving score, resource collection, CALVIN task success
& L0--L3, L7 \\
\midrule
\textbf{EWMBench} \citep{ewmbench2025}
& embodied robot-video benchmark
& Scene consistency, trajectory correctness, dynamics, semantics, diversity, and logical consistency in robot manipulation videos
& SceneC, HSD, nDTW, DYN, Diversity, BLEU, CLIP score, logical-error penalty
& L1--L3 \\
\midrule
\textbf{DreamGen Bench} \citep{dreamgen2025}
& benchmark for robot video world models used in policy learning
& Instruction following and physics alignment across RoboCasa and GR1 object/behavior/environment generalization settings
& GPT-4o, Qwen2.5-VL, human evaluation, VideoCon-Physics; IF, PA
& L2--L3 \\
\midrule
\textbf{WoW-World-Eval} \citep{wowworldeval2026}
& embodied world-model Turing test
& Video quality, instruction understanding, physical law, planning DAG quality, replay executability, OOD generalization, and human deception
& FVD/PSNR/SSIM/DINO/DreamSim; sequence match and execution quality; trajectory L2/DTW/FD; replay success; deceive-human ratio
& L0--L4, L7 \\
\midrule
\textbf{RBench} \citep{rbench2026}
& robot-oriented video generation benchmark
& Task correctness and embodiment-specific plausibility for common manipulation, long-horizon planning, collaboration, spatial relations, and visual reasoning
& VLM/LLM judging of PSS, TAC, RSS; programmatic motion amplitude and smoothness
& L0--L3 \\
\midrule
\textbf{WorldArena} \citep{worldarena2026}
& unified benchmark for perception and functional utility
& Open-loop video quality plus world model as data engine, policy evaluator, and action planner
& 16 video metrics; EWMScore; policy performance gain; correlation with simulator; planner success; human evaluation
& L0--L4, L6, L7 \\
\midrule
\textbf{RoboWM-Bench} \citep{robowmbench2026}
& benchmark for world models in robotic manipulation
& Whether generated human-hand or robot videos can be converted into executable actions that complete tasks in simulation
& Task-level and step-level success; real-to-sim consistency; video-to-action replay reliability
& L4, L7 \\
\midrule
\textbf{WorldScore} \citep{worldscore2025}
& world-generation benchmark
& Static and dynamic world generation under layout and camera control
& Camera control, object control, content alignment, 3D consistency, photo consistency, motion accuracy, motion smoothness, WorldScore
& L0--L3 \\
\midrule
\textbf{WorldModelBench} \citep{worldmodelbench2025}
& benchmark for judging video generators as world models
& Instruction following, commonsense, and physics adherence of generated videos across domains
& Human labels and trained VLM judge; instruction-following level, framewise/temporal quality, physics sub-scores, ELO
& L2--L3 \\
\midrule
\textbf{WorldPrediction} \citep{worldprediction2025}
& high-level world modeling and procedural planning benchmark
& Multiple-choice action or action-sequence selection from initial and final states
& Single-step world-modeling accuracy; multi-step procedural-planning accuracy
& L4, L7 \\
\midrule
\textbf{AutumnBench / WorldTest} \citep{autumnbench2025}
& environment-level query benchmark after exploration
& Masked frame prediction, change detection, and planning in text-based grid-world POMDPs
& MFP accuracy, change-detection score, planning success, aggregate score
& L1, L4, L7 \\
\midrule
\textbf{PBench} \citep{nvidia2025pbench}
& Physical-AI image-to-video benchmark
& Domain-specific physical and commonsense QA plus generic video quality
& Domain score via Qwen2.5-VL QA; VBench-style quality score; overall score
& L0--L3 \\
\midrule
\textbf{MVP} \citep{mvp2025}
& shortcut-resistant physical video QA
& Minimal-pair physical understanding across human-object, robot-object, intuitive-physics, and temporal-reasoning settings
& Paired minimal-pair accuracy
& L3 \\
\midrule
\textbf{IntPhys 2} \citep{intphys2_2025}
& intuitive-physics benchmark
& Possible vs.\ impossible events in complex synthetic scenes
& Overall and difficulty-split accuracy; pairwise and single-video evaluation
& L3 \\
\midrule
\textbf{CausalVQA} \citep{causalvqa2025}
& causal reasoning benchmark for video models
& Counterfactual, hypothetical, anticipation, planning, and descriptive reasoning on real egocentric videos
& Paired accuracy, unpaired accuracy, reasoning accuracy, difficulty splits, human baseline
& L3 \\
\bottomrule
\end{longtable}
\normalsize

The benchmark literature has moved beyond pure FID/FVD. Newer suites ask about instruction following, physical plausibility, trajectory correctness, executability, and policy-centric use \citep{worldsimbench2024,worldarena2026,robowmbench2026}. Even so, many benchmark suites place most of their evaluative weight on \emph{artifact quality} (e.g., the generated video, description, or QA answer) rather than on whether the model supports reliable policy evaluation or policy improvement. In objective-mismatch terms, these suites measure predictive or perceptual quality, which is known to be an unreliable proxy for control utility \citep{lambert2020objmismatch}.

\subsection{Environment-model and policy-optimization literature}

\scriptsize
\begin{longtable}{p{2.4cm}p{2.7cm}p{4.35cm}p{4.4cm}p{1.35cm}}
\caption{Environment-model and policy-optimization literature. These papers are closer to the original decision-making use of world models, but they vary in how directly they evaluate that use.}\label{tab:opt-survey}\\
\toprule
\textbf{Work} & \textbf{Primary use claim} & \textbf{What is actually evaluated} & \textbf{Representative metrics or outputs} & \textbf{Main levels} \\
\midrule
\endfirsthead
\toprule
\textbf{Work} & \textbf{Primary use claim} & \textbf{What is actually evaluated} & \textbf{Representative metrics or outputs} & \textbf{Main levels} \\
\midrule
\endhead
\textbf{WHALE} \citep{zhang2024whale}
& generalizable embodied decision world model
& Value estimation, video fidelity, uncertainty estimation, and downstream offline policy optimization under generalization shift
& Value-estimation quality, video fidelity, uncertainty quality, policy optimization gains
& L4--L7 \\
\midrule
\textbf{ACEM / GALILEO} \citep{chen2023acem}
& counterfactual environment-model learning
& Counterfactual prediction, off-policy evaluation, offline RL, and online decision making under behavior-policy bias
& Counterfactual prediction accuracy, OPE quality, policy-improvement performance
& L4, L6, L7 \\
\midrule
\textbf{ADM-v2} \citep{lin2026admv2}
& full-horizon dynamics model for offline learning
& Full-horizon rollout quality via off-policy evaluation and offline RL
& OPE reliability, full-horizon rollout performance, offline RL returns
& L4, L6, L7 \\
\midrule
\textbf{PCM} \citep{chen2024pcm}
& policy-conditioned environment model
& Value estimation, policy selection, and MPC under policy-distribution shift
& Value-gap reduction, policy-evaluation quality, MPC performance
& L4, L6, L7 \\
\midrule
\textbf{DayDreamer} \citep{wu2023daydreamer}
& world model for real-robot RL
& End-to-end policy learning on real robots; no standalone world-model benchmark
& Policy success, sample efficiency, robot learning time
& L7 \\
\midrule
\textbf{UniSim} \citep{yang2024unisim}
& interactive real-world simulator
& Video-generation quality plus downstream policy learning and synthetic-video-based captioning
& FID, FVD, IS, CLIP; Language Table success; CIDEr
& L0--L1, L7 \\
\midrule
\textbf{DiWA} \citep{diwa2025}
& diffusion-policy adaptation with world models
& Reward classifier quality, imagined PPO updates, and downstream policy adaptation
& Precision/Recall for reward; policy success; sample efficiency
& L5, L7 \\
\midrule
\textbf{World4RL} \citep{world4rl2025}
& diffusion world model for policy refinement
& Open-loop video quality plus manipulation policy refinement in simulation and real robots
& FID, FVD, LPIPS; success rate; interaction cost
& L0--L1, L7 \\
\midrule
\textbf{VLA-RFT} \citep{vlarft2025}
& world-simulator RL fine-tuning for VLAs
& Image prediction quality and downstream success/robustness on LIBERO
& MSE, PSNR, SSIM, LPIPS; success rate; perturbation success
& L1, L7 \\
\midrule
\textbf{ProphRL} \citep{prophrl2025}
& future-video world model with VLM reward
& Video prediction, optical-flow correctness, reward precision/recall, and policy success
& PSNR, SSIM, tSSIM, flow EPE/cosine; RM precision/recall/FPR; success
& L1, L5, L7 \\
\midrule
\textbf{WMPO} \citep{wmpo2026}
& world-model-based policy optimization for VLA models
& Policy optimization outcome, successful-trajectory length, and continual-learning performance
& Success rate, successful trajectory length, lifelong-learning success
& L7 \\
\midrule
\textbf{World-Env} \citep{rehearsevla2026}
& simulated post-training with a world model as virtual environment
& Video quality, reward-model quality, and policy success in simulation and real robots
& FID, FVD, PSNR, SSIM, LPIPS; RM accuracy/precision/recall/F1; success
& L1, L5, L7 \\
\midrule
\textbf{World-Gymnast} \citep{worldgymnast2026}
& VLA RL inside a video world model
& Mainly downstream policy performance under imagined RL
& Success rate on tabletop and real-robot tasks
& L7 \\
\midrule
\textbf{RISE} \citep{rise2026}
& self-improving robot policy with compositional world model
& Multi-view world-model quality, progress-value modeling, and real-robot policy success
& PSNR, SSIM, LPIPS, FVD, EPE; success and sub-step scores
& L1, L5, L7 \\
\midrule
\textbf{VLAW} \citep{vlaw2026}
& iterative co-improvement of VLA and world model
& Video quality, interaction-event correctness, reward quality, and policy success
& PSNR, SSIM, LPIPS, FID, FVD; TP/FN/TN/FP for interaction and reward; success
& L1, L4, L5, L7 \\
\midrule
\textbf{GigaBrain-0.5M} \citep{gigabrain2026}
& VLA trained with world-model-based RL
& Process reward/value prediction quality and downstream real-robot success
& Inference time, MAE, MSE, RMSE, Kendall's tau; success rate
& L5, L7 \\
\midrule
\textbf{WoVR} \citep{wovr2026}
& reliable simulator for VLA post-training
& Long-horizon video quality, speed, and downstream policy success
& LPIPS, FID, FVD, FloLPIPS, FPS; success rate
& L1, L7 \\
\midrule
\textbf{World-VLA-Loop} \citep{worldvlaloop2026}
& closed-loop co-training of world model and VLA policy
& World-model image quality, reward accuracy, and final success
& SSIM, PSNR, LPIPS, MSE; reward accuracy; success rate
& L1, L5, L7 \\
\midrule
\textbf{PlayWorld} \citep{playworld2026}
& world model learned from autonomous play
& Video quality, progress-reward modeling, failure-mode alignment, and policy success
& LPIPS, SSIM, PSNR, MSE; RM accuracy; failure-mode alignment; success
& L1, L5, L7 \\
\midrule
\textbf{VLA-MBPO} \citep{vlambpo2026}
& practical model-based RL for VLA models
& Image prediction, reward prediction, inference time, rollout-length ablation, and downstream success
& LPIPS, PSNR, SSIM; reward ACC/F1; inference time; success
& L1, L5, L7 \\
\bottomrule
\end{longtable}
\normalsize

This family lies closer to the decision-making end of the spectrum because it evaluates what policies achieve when the model is used for imagination, fine-tuning, or planning. Many papers in this family combine lower-level video diagnostics with final policy success, which leaves the contribution of the world model itself only partially isolated. The clearest exceptions are the papers that make counterfactual generalization, full-horizon OPE, policy shift, or uncertainty an explicit part of the evaluation, such as ACEM, PCM, ADM-v2, and WHALE \citep{chen2023acem,chen2024pcm,lin2026admv2,zhang2024whale}; we note these as exemplars of the evaluation \emph{style} we advocate, not as the only or best instances, and several policy-evaluation papers discussed below adopt a comparably decision-aligned style.

\subsection{Policy evaluation, executability, and synthetic-data literature}

\scriptsize
\begin{longtable}{p{2.45cm}p{2.55cm}p{4.35cm}p{4.45cm}p{1.35cm}}
\caption{Policy-evaluation, executability, and synthetic-data literature. These works move from artifact quality toward decision utility, often through pipeline-level metrics.}\label{tab:pe-survey}\\
\toprule
\textbf{Work} & \textbf{Primary use claim} & \textbf{What is actually evaluated} & \textbf{Representative metrics or outputs} & \textbf{Main levels} \\
\midrule
\endfirsthead
\toprule
\textbf{Work} & \textbf{Primary use claim} & \textbf{What is actually evaluated} & \textbf{Representative metrics or outputs} & \textbf{Main levels} \\
\midrule
\endhead
\textbf{WorldGym} \citep{worldgym2025}
& world model as environment for policy evaluation
& Qualitative rollout fidelity, qualitative action controllability, and correlation between model-evaluated and real policy success
& Pearson correlation; relative policy ranking
& L4, L6 \\
\midrule
\textbf{Vid2World} \citep{vid2world2026}
& interactive world model from video diffusion
& Video prediction quality and policy evaluation on robotic manipulation
& FVD, FID, SSIM, PSNR, LPIPS, DreamSim; simulated vs.\ real success
& L1, L6 \\
\midrule
\textbf{Scalable Policy Evaluation} \citep{tseng2025scalable}
& video world models as policy evaluators
& Video quality plus policy-value correlation and ranking fidelity
& PSNR, SSIM, FVD, latent L2; Pearson correlation; MMRV
& L1, L6 \\
\midrule
\textbf{Gemini/Veo Simulator} \citep{geminiveo2025}
& world simulator for policy evaluation and safety
& Nominal evaluation, OOD ranking, and safety red-teaming in a video simulator
& Pearson correlation, MMRV, qualitative safety findings
& L6 \\
\midrule
\textbf{dWorldEval} \citep{dworldeval2026}
& robotic policy evaluation via diffusion world model
& Action controllability, round-trip consistency, and policy-ranking fidelity
& \(\Delta\)-LPIPS, round-trip LPIPS, Pearson correlation, MMRV
& L4, L6 \\
\midrule
\textbf{IRASim} \citep{irasim2025}
& fine-grained world model for manipulation
& Video prediction, qualitative flexible controllability, policy evaluation, and model-based planning
& PSNR, SSIM, latent L2, FID, FVD; Pearson correlation; planning success
& L1, L4, L6, L7 \\
\midrule
\textbf{DreamDojo} \citep{dreamdojo2026}
& generalist robot world model
& Video prediction, human judgment of physics/action following, policy evaluation, and planning
& PSNR, SSIM, LPIPS; human physics/action-following; success, Pearson, MMRV
& L1, L3, L6, L7 \\
\midrule
\textbf{Kinema4D} \citep{kinema4d2026}
& 4D kinematic world model
& RGB rollout quality, geometry quality, and policy-evaluation quality
& PSNR, SSIM, latent L2, FID, FVD, LPIPS; Chamfer, F-score, temporal F-score; success-gap to ground truth
& L1, L3, L6 \\
\midrule
\textbf{Persistent Robot World Models} \citep{persistent2026}
& stabilized multi-step rollouts for policy evaluation
& Per-camera rollout quality, masked task-relevant metrics, human preference, and policy ranking
& SSIM, PSNR, LPIPS; temporal curves; masked metrics; 2AFC/ELO; Pearson, MMRV
& L1, L6 \\
\midrule
\textbf{DreMa} \citep{drema2025}
& synthetic imagination for imitation learning
& Object final-position accuracy and downstream imitation-learning gains
& Object final-position error; policy success
& L4, L7 \\
\midrule
\textbf{DreamGen} \citep{dreamgen2025}
& synthetic robot-video data engine
& VLM/human instruction following and physics alignment plus downstream policy learning and generalization
& IF, PA; downstream policy success
& L2, L3, L7 \\
\midrule
\textbf{Ctrl-World} \citep{ctrlworld2026}
& controllable generative world model
& Video quality, qualitative action controllability, policy evaluation correlation, and policy improvement with synthetic successful trajectories
& PSNR, SSIM, LPIPS, FID, FVD; evaluation correlation; success rate
& L1, L4, L6, L7 \\
\midrule
\textbf{RoboMaster} \citep{robomaster2026}
& embodied action planning from generated videos
& Bridge video quality, trajectory fidelity, user preference, action-planning success, and IDM action quality
& FVD, PSNR, SSIM, TrajError, user preference, planning success
& L1, L4, L7 \\
\midrule
\textbf{GigaWorld-0} \citep{gigaworld2025}
& world-model-based data engine
& Physical-AI/world-generation quality, filtering score, and action recovery from generated data
& PBench, DreamGen Bench, quality filtering, IDM action recovery
& L0--L4 \\
\midrule
\textbf{RoboVIP} \citep{robovip2026}
& multi-view video augmentation for manipulation
& Multi-view generation quality and downstream policy gains
& FID, FVD, LPIPS, MV-Mat; success rate
& L1, L7 \\
\midrule
\textbf{Interactive World Simulator} \citep{interactiveworldsim2026}
& interactive simulator for policy training and evaluation
& Video prediction, speed/stability, simulator-collected training data, and evaluation correlation
& MSE, LPIPS, FID, PSNR, SSIM, UIQI, FVD; FPS/stability; success; correlation
& L1, L6, L7 \\
\midrule
\textbf{VLP} \citep{vlp2024}
& video language planning
& Human judgment of long-horizon video-plan completion and downstream execution quality
& Human plan-completion rate; task reward/completion
& L2, L7 \\
\midrule
\textbf{Dreamitate} \citep{dreamitate2024}
& video generation for visuomotor policy learning
& Real-robot policy success using generated videos as supervision or guidance
& Success rate
& L7 \\
\midrule
\textbf{RoboDreamer} \citep{robodreamer2024}
& compositional world model for robot imagination
& Video generation quality, human task completion judgment, and execution success
& FVD; human completion judgment; RLBench success
& L0, L2, L7 \\
\midrule
\textbf{RoboEnvision} \citep{roboenvision2025}
& long-horizon robot video generation
& Long-horizon video quality and downstream policy-model task success
& LPIPS, SSIM, PSNR, FVD, CLIP score; success rate
& L1, L2, L7 \\
\midrule
\textbf{Genie Envisioner} \citep{genieenvisioner2026}
& unified platform for manipulation
& Real-robot action-model success and EWMBench-style scene/motion/semantic world-model evaluation
& Success rate; SceneC, HSD, nDTW, DYN, Diversity, BLEU, CLIP, Logics
& L1, L2, L3, L7 \\
\midrule
\textbf{EVA (model)} \citep{eva_model2026}
& executable video world model via IDM rewards
& Human judgment of kinematic plausibility, interaction plausibility, instruction adherence, and execution success in sim and real robots
& Human ratings; simulator success; real-robot success; IDM task success
& L2, L3, L7 \\
\bottomrule
\end{longtable}
\normalsize

This family is central for a decision-making-centric paper because it contains some of the strongest current attempts to evaluate world models as \emph{functional} objects rather than as generators. The move is away from asking only ``Does the video look plausible?'' and toward asking ``Does the model rank policies correctly?'', ``Can it support planning?'', or ``Can generated futures be executed?'' Many of these are still \emph{pipeline metrics}: executability depends on inverse dynamics or retargeting; synthetic-data value depends on the downstream learner; policy-evaluation correlation depends on reward checkers and the policy set being ranked.

\subsection{Latent and foundation-model literature}

\scriptsize
\begin{longtable}{p{2.55cm}p{2.55cm}p{4.3cm}p{4.55cm}p{1.25cm}}
\caption{Latent and foundation-model literature. ``repr.'' denotes cross-cutting state-abstraction or representation diagnostics that do not map cleanly to a single rung of the ladder but that matter chiefly because they support higher-level decision claims.}\label{tab:latent-survey}\\
\toprule
\textbf{Work} & \textbf{Primary use claim} & \textbf{What is actually evaluated} & \textbf{Representative metrics or outputs} & \textbf{Main levels} \\
\midrule
\endfirsthead
\toprule
\textbf{Work} & \textbf{Primary use claim} & \textbf{What is actually evaluated} & \textbf{Representative metrics or outputs} & \textbf{Main levels} \\
\midrule
\endhead
\textbf{Cosmos Predict 2.5} \citep{cosmos2025}
& diffusion-based world foundation model
& Physical-AI domain competence, generic video quality, multi-view robot geometry, action-conditioned robot prediction, and DreamGen-Bench performance
& Domain score, quality score, TransErr, RotErr, Sampson error, PSNR, SSIM, latent L2, FVD
& L0--L4 \\
\midrule
\textbf{ABot-PhysWorld} \citep{abotphysworld2026}
& interactive world foundation model for manipulation
& PBench/EZS-Bench world-generation quality and action-to-video trajectory consistency
& Domain/Robot score, VBench-style quality metrics, trajectory consistency
& L0--L4 \\
\midrule
\textbf{EA-WM} \citep{eawm2026}
& event-aware generative world model
& Interaction quality, trajectory accuracy, depth accuracy, perspectivity, instruction following, semantic alignment, action following, and action recoverability
& VLM interaction and perspective scores; trajectory NDTW; depth accuracy; KVAF translation/rotation/gripper error
& L2--L4 \\
\midrule
\textbf{V-JEPA 2} \citep{assran2025vjepa2}
& latent predictive world model for understanding, prediction, and planning
& Qualitative decoded rollouts, robot planning, action anticipation, classification, and video QA
& Goal distance, success rate, planning time, Top-1 accuracy, Recall@5, VQA accuracy
& L7 + repr. \\
\midrule
\textbf{V-JEPA 2.1} \citep{vjepa21_2026}
& dense-feature latent predictive model
& Robot planning, navigation, dense prediction, action anticipation, classification, and video QA
& Success rate, ATE/RTE, RMSE, mIoU, J\&F, Recall@5, Top-1 accuracy
& L7 + repr. \\
\midrule
\textbf{LeWorldModel} \citep{leworldmodel2026}
& latent JEPA-style world model from pixels
& Latent-space MPC planning, latent physical probing, and detection of physically implausible latent trajectories
& Planning success, MSE, Pearson correlation, anomaly separation
& L5, L7 + repr. \\
\midrule
\textbf{Implicit World Model Evaluation} \citep{vafa2024implicit}
& formal evaluation of the state-transition structure learned by generative models
& Whether the model merges histories that correspond to the same true state and separates histories with different future possibilities
& Sequence compression and sequence distinction
& repr. \\
\bottomrule
\end{longtable}
\normalsize

These papers make two points. First, a world model need not be a pixel generator: V-JEPA 2, V-JEPA 2.1, and LeWorldModel are evaluated mainly through planning, probing, and downstream utility \citep{assran2025vjepa2,vjepa21_2026,leworldmodel2026}. This echoes, informally, the value-equivalence intuition that a model can be useful by capturing the value-relevant subspace without modeling pixels \citep{grimm2020value}, though these latent models typically do not define an explicit reward/value set against which equivalence is formally stated. Second, strong next-step or surface-level prediction can be misleading about state abstraction: Vafa et al.\ show that a model may appear strong under standard probes while still failing to recover coherent transition structure \citep{vafa2024implicit}. A decision world model may ultimately depend more on the right \emph{state-transition abstraction} than on the right pixels.

\subsection{What is actually being evaluated?}

Taken together, Tables~\ref{tab:bench-survey}--\ref{tab:latent-survey} suggest that the literature is not evaluating a single object.

\paragraph{First, there are multiple evaluation cultures.}
Benchmark papers such as EVA-Bench, EWMBench, WorldModelBench, PBench, RBench, and WorldScore mainly evaluate the \emph{generated artifact}: realism, semantics, or physics \citep{eva2024,ewmbench2025,worldmodelbench2025,nvidia2025pbench,rbench2026,worldscore2025}. Policy-evaluation papers such as WorldGym, Scalable Policy Evaluation, dWorldEval, and DreamDojo evaluate the model as a \emph{surrogate evaluator of policies} \citep{worldgym2025,tseng2025scalable,dworldeval2026,dreamdojo2026}. Optimization papers evaluate the model as part of an \emph{end-to-end control-improvement pipeline} \citep{wu2023daydreamer,world4rl2025,vlarft2025,wmpo2026}. Latent papers evaluate the model as a \emph{planning-relevant representation} \citep{assran2025vjepa2,vjepa21_2026,leworldmodel2026}. Synthetic-data and executability papers evaluate the model as a \emph{data engine or executable planner} \citep{dreamgen2025,robowmbench2026,robomaster2026,eva_model2026}.

\paragraph{Second, the distribution of evidence is skewed toward lower levels, read directly off the survey.}
Tables~\ref{tab:opt-survey} and~\ref{tab:pe-survey} together list roughly forty environment-model, optimization, policy-evaluation, executability, and synthetic-data works, the subset of the survey that most plausibly carries decision-making claims. Reading the recorded metrics under the rule of Section~3, three patterns stand out. (i) Open-loop reconstruction (L1) and final policy success (L7) are the two most common rungs, and a large fraction of works report \emph{only} this pair as their world-model evidence, leaving the model's specific contribution entangled with the optimizer, reward model, and data pipeline. (ii) Fixed-policy ranking agreement (L6) appears in only about a dozen works, almost all of them in the dedicated policy-evaluation cluster (e.g., WorldGym, Scalable Policy Evaluation, dWorldEval, DreamDojo, Persistent, Ctrl-World, IRASim). (iii) Most strikingly, we find essentially no work in these tables that reports an \emph{exploitability gap}, a measured discrepancy between model-predicted value and real value for policies or action sequences \emph{optimized against the model} (the \(\mathrm{XGap}\) of Section~5). A handful report related diagnostics, such as a success-gap to ground truth for fixed rollouts (Kinema4D) or uncertainty quality (WHALE), but the specific failure mode that model-based optimization is known to provoke is, by the evidence of our own survey, almost never measured. This absence is the concrete form of the claim that the field has re-encountered objective mismatch without adopting the diagnostics developed for it. We present these counts as a reading of the metrics recorded in our tables rather than as a full audit of every paper's appendix; the qualitative conclusion is robust to small recounting differences.

\paragraph{Third, the main fault line is not simply ``video vs.\ RL.''}
The deeper distinction is between evaluating \emph{the generated artifact} and evaluating \emph{the decisions enabled by the model}. Many VLA/RL papers report policy success but diagnose the world model mainly through L1-style reconstruction metrics. Many benchmark papers add semantics and physics but remain artifact-centered. Policy-evaluation and executability papers move closer to decision utility, but often through black-box or pipeline-level metrics.

\paragraph{Fourth, pixels are neither necessary nor sufficient, and the divergence is observed, not merely possible.}
V-JEPA 2, V-JEPA 2.1, and LeWorldModel show that a model can be useful for planning without photorealistic reconstruction \citep{assran2025vjepa2,vjepa21_2026,leworldmodel2026}. The converse, high visual quality with low functional utility, is exactly what the two benchmarks designed to separate these axes report. WorldArena pairs sixteen video-quality metrics with functional measures (data-engine gain, policy-evaluation correlation, planner success) and finds that perceptual ranking and functional ranking of world models do not coincide \citep{worldarena2026}; RoboWM-Bench similarly evaluates whether generated videos can be \emph{executed} into task success and reports that video quality is not a reliable predictor of executability \citep{robowmbench2026}. We emphasize that these are published findings on real model sets, not a thought experiment: the artifact-vs-utility divergence the objective-mismatch literature predicts is already being measured in the generative-world-model setting, wherever a benchmark bothers to report both axes. The illustrative example in Section~5 is included only to make the mechanism intuitive, not to establish that the divergence occurs.

\subsection{The world-model claim}

Different papers use the same term, ``world model,'' while implicitly making different claims:

\begin{enumerate}[leftmargin=1.5em]
    \item \textbf{Future-video claim:} the model predicts plausible or realistic future observations.
    \item \textbf{Policy-evaluation claim:} the model can estimate or rank the performance of candidate policies.
    \item \textbf{Policy-optimization claim:} using the model inside a planner, optimizer, or RL loop improves policies in the real environment.
    \item \textbf{Planning/executability claim:} the model can help choose or recover action sequences that succeed in the real environment.
    \item \textbf{Synthetic-data claim:} model-generated samples improve downstream learning.
    \item \textbf{Representation claim:} the model's latent state preserves task-relevant transition structure sufficient for prediction and control.
\end{enumerate}

These claims are not interchangeable. Evidence for one of them is not automatically evidence for another.

\begin{table*}[t]
\centering
\small
\begin{tabularx}{\textwidth}{p{2.8cm} X X X}
\toprule
\textbf{Claim type} & \textbf{What stronger supporting evidence would typically include} & \textbf{Common weaker evidence sometimes substituted} & \textbf{Why the substitution can be misleading} \\
\midrule
Future-video / world-generation claim &
L0--L3 evidence: visual plausibility, logged-future fidelity, semantic alignment, physical plausibility &
None; these may be appropriate if this is the actual claim &
The problem begins only when this evidence is later treated as if it also established policy evaluation or optimization utility. \\
\midrule
Policy-evaluation claim &
Closed-loop rollouts of fixed policies with value/ranking agreement, ideally supported by L4--L5 diagnostics &
FVD/PSNR, human preference, or instruction-following scores alone &
A model can generate plausible videos while misranking policies or misestimating success. \\
\midrule
Policy-optimization claim &
Real or trusted-sim policy lift under fixed budgets, plus exploitability tests and L4--L6 diagnostics &
Open-loop image/video metrics, or a single downstream success number without decomposition &
Final success entangles the world model with reward models, filters, optimizers, and data curation; open-loop metrics do not test interventional usefulness. \\
\midrule
Planning / executability claim &
Task success under model-based planning or action recovery, ideally with action-intervention diagnostics &
Instruction-following videos or human preference for plausible plans &
A plausible-looking plan can still be dynamically wrong or non-executable. \\
\midrule
Synthetic-data claim &
Controlled downstream learning gains under matched training budgets and learners &
Video aesthetics, FVD, or VLM preference over generated data &
Visually clean data need not contain the right task-relevant variation for learning. \\
\midrule
Representation claim &
Planning utility, physical probes, or state-abstraction diagnostics such as sequence compression/distinction &
Decoder quality or linear probes alone &
A model can reconstruct well or support shallow probing while still lacking the correct transition structure. \\
\bottomrule
\end{tabularx}
\caption{A recurring issue in the literature is claim/evidence mismatch: lower-level evidence is sometimes taken to support stronger world-model claims than it can justify. This is the objective-mismatch problem \citep{lambert2020objmismatch} restated at the level of evaluation.}
\label{tab:claim-mismatch}
\end{table*}

This observation can be made concrete in the recent literature. Benchmarks such as PBench, WorldModelBench, WorldScore, and RBench are informative if the claim is embodied world generation or physical-video quality, but they do not by themselves establish policy-evaluation or policy-optimization utility \citep{nvidia2025pbench,worldmodelbench2025,worldscore2025,rbench2026}. Many RL/VLA papers such as World4RL, VLA-RFT, World-Env, RISE, WoVR, and VLA-MBPO evaluate final policy success, which is important, but often diagnose the world model itself mainly through L1-style image or video metrics \citep{world4rl2025,vlarft2025,rehearsevla2026,rise2026,wovr2026,vlambpo2026}. Policy-evaluation papers such as WorldGym, Scalable Policy Evaluation, Vid2World, dWorldEval, DreamDojo, and Persistent Robot World Models are closer to the decision-making use, yet they mostly evaluate fixed-policy ranking rather than optimizer interaction or exploitability \citep{worldgym2025,tseng2025scalable,vid2world2026,dworldeval2026,dreamdojo2026,persistent2026}. Executability papers such as RoboWM-Bench, RoboMaster, and EVA test a stronger bridge from video to control, but their performance also depends on inverse dynamics, retargeting, simulators, and success checkers \citep{robowmbench2026,robomaster2026,eva_model2026}. Latent papers show the converse point: planning utility can exist without high-quality pixels \citep{assran2025vjepa2,leworldmodel2026,vafa2024implicit}.

The conclusion is not that the literature has been ``doing it wrong.'' Rather, the field has been evaluating several different things under one name, and has often re-discovered the objective-mismatch problem without using the existing decision-aware vocabulary to address it. The next section organizes the evaluation targets explicitly.

\section{World-Model Evaluation Ladder}

The survey suggests an L0--L7 ladder of evaluation targets. The ladder is \emph{descriptive}, because it summarizes what the literature already measures, and partly \emph{normative}, because the levels answer increasingly strong questions about whether a model supports embodied decision-making. The point is not that lower levels are useless; it is that lower levels generally answer weaker questions about decision use.

\paragraph{The ladder spans several axes, not one.}
We caution at the outset that ``ladder'' is a simplification. The L0--L7 ordering actually moves along several partly independent axes: artifact-quality versus decision-utility (L0--L3 vs.\ L4--L7), observational versus interventional (L0--L3 vs.\ L4 onward), open-loop versus closed-loop (L1 vs.\ L6--L7), and fixed-policy versus optimization-induced (L6 vs.\ L7). These axes are correlated but not collinear. We retain the linear presentation only because it tracks, roughly, the strength of the decision-making question being asked; it should be read as an evidential hierarchy, not as a single scalar scale, and the level assignments in Tables~\ref{tab:bench-survey}--\ref{tab:latent-survey} are sets of levels precisely because most papers occupy several axes at once.

\begin{table*}[h]
\centering
\small
\begin{tabularx}{\textwidth}{c p{2.5cm} X X X}
\toprule
\textbf{Level} & \textbf{Name} & \textbf{Core question} & \textbf{Typical metrics} & \textbf{Representative works} \\
\midrule
L0 & Visual plausibility &
Does the output look like a realistic image or video? &
FID, FVD, aesthetics, image quality, human preference, VBench-style scores &
UniSim, WorldScore, PBench, Cosmos Predict 2.5, ABot-PhysWorld \\
\midrule
L1 & Logged-future prediction &
Does the predicted future match held-out trajectories from the behavior distribution? &
MSE, PSNR, SSIM, LPIPS, DreamSim, latent L2, tSSIM, optical-flow error &
World4RL, VLA-RFT, ProphRL, IRASim, Kinema4D, Persistent \\
\midrule
L2 & Semantic alignment &
Does the rollout match the instruction, task, and scene semantics? &
CLIPScore, caption metrics, VLM/LLM judges, task-completion labels &
EVA-Bench, DreamGen Bench, RBench, WorldArena, EA-WM \\
\midrule
L3 & Physical plausibility &
Does the rollout obey intuitive physical and geometric constraints? &
Physics QA, VLM physics scores, object permanence, depth, contact, trajectory, 3D consistency &
WorldModelBench, PBench, MVP, IntPhys 2, CausalVQA, WorldArena \\
\midrule
L4 & Action controllability / interventional fidelity &
Does changing actions produce the correct task-relevant changes? &
Action-effect tests, trajectory error, \(\Delta\)-LPIPS, round-trip consistency, interventional OPE diagnostics &
ACEM, PCM, WHALE, dWorldEval, WorldGym, IRASim, RoboMaster \\
\midrule
L5 & Reward, value, and outcome fidelity &
Does the model predict success, reward, progress, or value accurately enough for decision making? &
Reward accuracy, precision/recall/F1, calibration, value error, Kendall's tau, progress ranking &
DiWA, ProphRL, RISE, VLAW, GigaBrain, PlayWorld, VLA-MBPO \\
\midrule
L6 & Policy evaluation and ranking &
Does model-based evaluation agree with real/simulator policy performance? &
Pearson/Spearman correlation, pairwise ranking accuracy, MMRV, calibration &
WorldGym, Vid2World, Scalable Policy Evaluation, dWorldEval, DreamDojo \\
\midrule
L7 & Policy optimization / planning utility &
Does using the model improve decisions? &
Policy lift, optimization regret, sample efficiency, planning success, safe improvement, exploitability gap &
DayDreamer, ADM-v2, WMPO, DreamGen, VLP, EVA, V-JEPA 2 \\
\bottomrule
\end{tabularx}
\caption{The L0--L7 ladder. Levels are ordered by the strength of the question they answer about world-model usefulness for embodied decision making, but they span several orthogonal axes (artifact/decision, observational/interventional, open-loop/closed-loop, fixed-policy/optimization).}
\label{tab:ladder}
\end{table*}

The levels are not mutually exclusive, and they are not strictly monotone in practice. A latent model might score modestly on L0 or L1 and still be useful at L7; a video model might score highly at L0--L3 and still be weak at L6. The ladder is therefore best read as a hierarchy of increasingly direct evidence for decision-making claims, not as a single linear scorecard.

\textbf{L0 (Visual plausibility)} asks whether the output \emph{looks} like a plausible image or video. This is the dominant level in video-generation evaluations, where metrics such as FID, FVD, aesthetics, image quality, and human preference are natural first checks \citep{yang2024unisim,worldscore2025,nvidia2025pbench,cosmos2025,abotphysworld2026}. L0 is useful because obvious visual failures often indicate model collapse, temporal artifacts, or poor conditioning, and it matters when humans inspect rollouts or when generated data must pass a basic realism filter. But L0 is only indirect evidence of decision usefulness: a model can attain good L0 by producing smooth videos, copying backgrounds, or generating semantically plausible but action-insensitive futures. L0 therefore diagnoses surface quality, not whether the model gets the \emph{consequences of chosen actions} right.

\textbf{L1 (Logged-future prediction)} asks whether a predicted future matches a held-out future from the behavior distribution. This is the standard open-loop world-model setting of MSE, PSNR, SSIM, LPIPS, DreamSim, latent L2, temporal SSIM, or optical-flow accuracy \citep{world4rl2025,vlarft2025,prophrl2025,irasim2025,kinema4d2026,persistent2026}. L1 is a useful sanity check: a model that cannot predict held-out futures at all is less likely to support higher-level use. However, L1 remains observational and is exactly the quantity the objective-mismatch literature found to be weakly correlated with control performance \citep{lambert2020objmismatch}: it measures whether the model matches \emph{what happened}, not \emph{what would happen under different actions or policies}, and in stochastic settings it can penalize plausible alternative futures. A model can be mediocre at L1 but useful for planning if it preserves reward-relevant structure; conversely, it can be strong at L1 yet fail under policy shift.

\textbf{L2 (Semantic alignment)} evaluates whether the rollout matches the instruction, task, objects, and scene semantics. This level appears in EVA-Bench, DreamGen Bench, RBench, WorldArena, WoW-World-Eval, and EA-WM \citep{eva2024,dreamgen2025,rbench2026,worldarena2026,wowworldeval2026,eawm2026}, via CLIPScore, caption overlap, VLM/LLM judges, and task-completion labels. L2 is often important for language-conditioned systems, because a model that misunderstands the task is unlikely to support meaningful downstream use. Still, L2 remains artifact-centered: VLM judges and semantic similarity scores can reward plausible narratives rather than faithful dynamics, and a video may appear to follow the instruction while encoding the wrong object contact, force direction, or success state.

\textbf{L3 (Physical plausibility)} asks whether the rollout obeys intuitive physics and geometric consistency: object permanence, continuity, gravity, non-penetration, contact, depth, or trajectory coherence. This level appears in WorldModelBench, PBench, WorldArena, MVP, IntPhys 2, CausalVQA, and EA-WM \citep{worldmodelbench2025,nvidia2025pbench,worldarena2026,mvp2025,intphys2_2025,causalvqa2025,eawm2026}. Relative to L0 and L2, L3 is a meaningful advance because it penalizes physically impossible or causally incoherent artifacts. Yet L3 is still not enough for decision use unless it is tied to interventions: many physical benchmarks ask whether a video looks physically plausible or whether a model answers a physical question correctly, which does not directly test whether the model predicts the consequences of \emph{the agent's} chosen actions in the regions of state-action space that matter for policy optimization.

\textbf{L4 (Action controllability and interventional fidelity)}
asks whether changing the action causes the \emph{correct task-relevant change}. This is explicit in ACEM, PCM, WHALE, dWorldEval, WorldGym, IRASim, and RoboMaster \citep{chen2023acem,chen2024pcm,zhang2024whale,dworldeval2026,worldgym2025,irasim2025,robomaster2026}. Typical evidence includes action-effect tests, end-effector or object trajectory accuracy, \(\Delta\)-LPIPS, round-trip consistency, or interventional OPE diagnostics. L4 is the first level that clearly distinguishes a decision world model from a future-video prior, and for embodied decision-making it is, in our view, a strong candidate for the minimal genuinely interventional requirement. If a model appears insensitive to actions or produces the wrong action-dependent changes, then strong L0--L3 scores provide limited reassurance about decision usefulness.

\textbf{L5 (Reward, value, and outcome fidelity)} asks whether the model predicts success, reward, progress, constraint violation, or value accurately enough for decision making. This is explicit in DiWA, ProphRL, RISE, VLAW, GigaBrain, PlayWorld, and VLA-MBPO \citep{diwa2025,prophrl2025,rise2026,vlaw2026,gigabrain2026,playworld2026,vlambpo2026}, via reward accuracy, precision/recall/F1, success-probability calibration, value error, or Kendall's tau for progress ranking. L5 matters because a visually imperfect model can still be useful if it preserves the variables that determine reward, whereas a photorealistic simulator that hallucinates success can be risky for policy optimization. This is, informally, the value-equivalence intuition made measurable, and a reason to treat reward and value as first-class evaluation targets rather than downstream afterthoughts \citep{grimm2020value}.

\textbf{L6 (Policy evaluation and ranking)} asks whether model-based evaluation agrees with real or simulator policy performance. This is the focus of WorldGym, Vid2World, Scalable Policy Evaluation, dWorldEval, DreamDojo, Gemini/Veo, and Persistent Robot World Models \citep{worldgym2025,vid2world2026,tseng2025scalable,dworldeval2026,dreamdojo2026,geminiveo2025,persistent2026}, via Pearson or Spearman correlation, pairwise ranking accuracy, and MMRV. This is one of the stronger \emph{fixed-policy} criteria in the current literature: it directly tests whether the model preserves the ordering of policies, which is often more decision-relevant than image similarity. But L6 is still weaker than L7, because an optimizer can drive the policy into parts of the space that were never tested by the fixed policy set.

\textbf{L7 (Policy optimization and planning utility)} asks the pragmatic question: does using the model improve decisions? This includes model-based planning, model-based RL, executable video planning, and synthetic-data-driven gains. DayDreamer, ADM-v2, WMPO, DreamGen, VLP, EVA, and V-JEPA 2 are representative \citep{wu2023daydreamer,lin2026admv2,wmpo2026,dreamgen2025,vlp2024,eva_model2026,assran2025vjepa2}. L7 provides the most direct evidence for world models claimed for embodied decision-making. At the same time, L7 is \emph{entangled}: it depends not only on the world model, but also on the reward model, the optimizer, the rollout horizon, uncertainty handling, and the surrounding data pipeline. This is why L7 is usually easier to interpret when paired with L4--L6 decompositions rather than treated as a standalone number.

The ladder suggests a useful distinction. \textbf{L0--L3 are best interpreted as diagnostic levels}: they evaluate the generated artifact and are useful and often necessary, especially for video-based interfaces. \textbf{L4 can be viewed as an interventional threshold}: it asks whether the model responds correctly to actions. \textbf{L5--L7 provide the most direct evidence of decision use}: they test whether the model preserves outcomes, ranks policies, and improves decisions. State-abstraction and latent-representation diagnostics such as \citep{vafa2024implicit} cut across the ladder rather than forming an extra rung; they matter chiefly because they help explain success or failure at L4--L7. For decision-making claims, lower-level success is not a reliable substitute for higher-level evidence, even though lower-level metrics remain useful in practice and may correlate with higher-level performance in some domains.

\section{A Decision-Making-Centric Evaluation Framework}

The ladder clarifies what kinds of evidence exist. We now turn from description to proposal. The points below are recommendations for \emph{models whose stated purpose is embodied decision-making}, not universal requirements that every world-model paper must satisfy.

\paragraph{A declared decision contract clarifies evaluation.}
There is no single use-independent world-model score; the value-equivalence principle makes this precise, since model fidelity is only defined relative to a set of policies and value functions \citep{grimm2020value,grimm2021proper}. Evaluation becomes easier to interpret once a \emph{decision contract} is declared:
\begin{quote}
\emph{This model is a world model for task family \(\T\), policy class \(\PiSet\), action interface \(\A\), horizon \(H\), and decision use \(U\).}
\end{quote}
Without \(\T\), \(\PiSet\), \(H\), and \(U\), it is difficult to know whether FVD, VLM-based physics QA, policy ranking, or final task success is the most relevant primary metric. This explains a large part of the survey: many benchmark papers are reasonable once interpreted as L0--L3 contracts, and the difficulty arises only when those results are generalized to stronger L6--L7 claims without additional evidence.

\paragraph{Interventional action fidelity is often the first distinguishing requirement.}
A passive video predictor estimates likely futures under the data distribution. A decision world model is more compelling when it can answer interventional queries, distinguishing
\[
    \Prob_{\Env}(\obs_{t+1:t+H} \mid \hist_t)
    \quad\text{from}\quad
    \Prob_{\Env}(\obs_{t+1:t+H} \mid \hist_t, do(\act_{t:t+H-1})).
\]
The second object is the one needed for planning and policy optimization, and (per Section~2.6) is identifiable from logs only under no-unobserved-confounding and coverage assumptions, and only when the action interface is directly controlled rather than inferred from generated video; otherwise it must be measured through executed interventions. For a task-relevant feature map \(\Phi\), define the interventional prediction error
\[
    \mathrm{IPE}_{H,\Phi}(\Model)
    =
    \E_{(\hist,\act_{0:H-1}) \sim \mathcal{Q}}
    \left[
        d_{\Phi}
        \left(
            \Phi(\widehat{\traj}_{1:H}),
            \Phi(\traj_{1:H})
        \right)
    \right],
\]
where \(\widehat{\traj}\sim\Model(\cdot\mid\hist,\act_{0:H-1})\) and \(\traj\sim\Env(\cdot\mid\hist,do(\act_{0:H-1}))\). The feature map \(\Phi\) may include object poses, end-effector states, contact events, success predicates, safety constraints, or latent task variables, depending on the domain. A complementary action-effect metric compares two actions from the same history:
\[
    \Delta_{\Env}
    =
    d_{\Phi}
    \left(
        \Phi(\traj^{\act}),
        \Phi(\traj^{\act'})
    \right),
    \qquad
    \Delta_{\Model}
    =
    d_{\Phi}
    \left(
        \Phi(\widehat{\traj}^{\act}),
        \Phi(\widehat{\traj}^{\act'})
    \right).
\]
A model that preserves the magnitude and ordering of action-induced differences offers stronger evidence of decision relevance than one that merely predicts plausible futures.

\paragraph{Policy-induced distribution shift is usually worth including.}
Behavior-policy data and target-policy rollouts generally come from different state-action distributions. Let \(d_{\Env}^{\pi}(\state,\act)\) denote the discounted occupancy measure of policy \(\pi\) in the real environment, and \(d_{\Model}^{\pi}\) the analogous occupancy under the model. A model may have low error under \(d_{\Env}^{\mu}\), where \(\mu\) is the behavior policy, but high error under \(d_{\Env}^{\pi}\), where \(\pi\) is the target or optimized policy. This is why policy-conditioned models, counterfactual environment-model learning, and full-horizon dynamics models are relevant here \citep{chen2023acem,chen2024pcm,lin2026admv2}. A decision-centric benchmark is more informative when it includes target policies that differ from the data-collection policies, ideally including policies produced by the model-based optimizer itself.

\paragraph{For many decision uses, full-horizon outcome fidelity is more informative than short-horizon reconstruction.}
One-step or short-horizon accuracy can be misleading: small local errors may compound, while visually large errors may be irrelevant to reward. For many decision uses, it is therefore more informative to evaluate the world model in terms of full-horizon task-relevant outcomes,
\[
    \Jhat_{\Model}(\pi)
    =
    \E_{\widehat{\traj}\sim \Model,\pi}
    \left[
        \sum_{t=0}^{H-1}\gamma^t \widehat{r}_t
    \right],
\]
and, for a policy set \(\PiSet_{\mathrm{eval}}\), the full-horizon value error
\[
    \mathrm{FVE}(\Model,\PiSet_{\mathrm{eval}})
    =
    \frac{1}{|\PiSet_{\mathrm{eval}}|}
    \sum_{\pi\in\PiSet_{\mathrm{eval}}}
    \left|
        \Jhat_{\Model}(\pi)-\J_{\Env}(\pi)
    \right|.
\]
For sparse-success tasks, success-probability calibration is also useful:
\[
    \widehat{p}_{\Model}(\pi)
    =
    \Prob_{\widehat{\traj}\sim \Model,\pi}
    \left[
        \widehat{\traj}\ \mathrm{succeeds}
    \right],
    \qquad
    p_{\Env}(\pi)
    =
    \Prob_{\traj\sim \Env,\pi}
    \left[
        \traj\ \mathrm{succeeds}
    \right].
\]
When the model is used to compare or optimize policies, it is often more important that it preserve what the policy is trying to optimize than that it preserve every visual detail.

\paragraph{Closed-loop rollout and occupancy fidelity can be more informative than teacher-forced prediction.}
A world model used by a policy is often more meaningfully evaluated \emph{closed-loop}: the policy acts on model-generated histories, not only on teacher-forced ground-truth prefixes. One useful target is occupancy mismatch,
\[
    D_{\mathrm{occ}}(\Model,\pi)
    =
    \sum_{t=0}^{H}
    \gamma^t
    D\!\left(
        d_{\Env,t}^{\pi},
        d_{\Model,t}^{\pi}
    \right),
\]
where \(D\) may be MMD, Wasserstein distance, KL divergence, total variation, or a task-specific discrepancy in feature space. A world model that remains stable only while conditioned on ground-truth context may still be useful for some applications, but it offers weaker evidence as a closed-loop simulator.

\paragraph{Policy ranking can often be measured directly.}
For policy evaluation, exact value may matter less than choosing the better policy. For \(\PiSet_{\mathrm{eval}}=\{\pi_1,\ldots,\pi_n\}\), pairwise ranking accuracy is
\[
    \mathrm{PRA}
    =
    \frac{1}{n(n-1)}
    \sum_{i\neq j}
    \mathbf{1}
    \left[
        \left(
            \Jhat_{\Model}(\pi_i)-\Jhat_{\Model}(\pi_j)
        \right)
        \left(
            \J_{\Env}(\pi_i)-\J_{\Env}(\pi_j)
        \right)
        >0
    \right],
\]
and mean maximum rank violation is
\[
    \mathrm{MMRV}
    =
    \frac{1}{n}
    \sum_{i=1}^{n}
    \max_{j:\J_{\Env}(\pi_i)>\J_{\Env}(\pi_j)}
    \left[
        \rank_{\Model}(\pi_i)-\rank_{\Model}(\pi_j)
    \right]_+.
\]
These are the kinds of quantities emphasized in the emerging policy-evaluation literature \citep{worldgym2025,tseng2025scalable,dworldeval2026}; we aggregate rather than originate them.

\paragraph{Optimization utility is easier to interpret alongside exploitability tests.}
If a fixed optimizer \(\A\) uses the world model to produce \(\widehat{\pi}_{\Model} = \A(\Model,\D,B)\), then the primary system-level quantity is policy lift,
\[
    \mathrm{Lift}
    =
    \J_{\Env}(\widehat{\pi}_{\Model})
    -
    \J_{\Env}(\pi_{\mathrm{base}}),
\]
and, when an oracle or strong reference \(\pi^\star\) is available, optimization regret \(\mathrm{OptRegret}=\J_{\Env}(\pi^\star)-\J_{\Env}(\widehat{\pi}_{\Model})\). A complementary exploitability metric is
\[
    \mathrm{XGap}
    =
    \Jhat_{\Model}(\widehat{\pi}_{\Model})
    -
    \J_{\Env}(\widehat{\pi}_{\Model}).
\]
A large positive exploitability gap indicates that the optimizer found trajectories that look good to the model but fail in the environment. This is the model-exploitation failure mode long noted in model-based RL \citep{janner2019mbpo,lambert2020objmismatch}; we restate it here only as an evaluation quantity, because, as the count in Section~3.6 shows, the recent generative-world-model literature essentially never reports it.

\paragraph{Uncertainty and abstention can affect evaluation.}
A world model used for optimization is easier to trust if it can indicate when it is unreliable. Let \(u_{\Model}(\hist,\act_{0:H-1})\) denote an uncertainty score. A useful uncertainty measure is ideally calibrated with respect to outcome or value error:
\[
    \Prob
    \left(
        \left|
            \Jhat_{\Model}(\pi)-\J_{\Env}(\pi)
        \right|
        \leq \epsilon
        \;\middle|\;
        u_{\Model}(\pi)\leq \alpha
    \right)
    \approx
    1-\delta.
\]
In practice, this can be reported through risk-coverage curves, error-vs.-uncertainty correlation, abstention performance, pessimistic-planning performance, Brier score, and \(\ECE\). This matters because offline or model-based optimizers may actively seek overconfident errors; WHALE's emphasis on generalization and uncertainty is therefore relevant for decision-centric evaluation rather than a side issue \citep{zhang2024whale}.

\paragraph{Estimation caveats.}
The quantities above are targets, not free measurements, and several are statistically demanding. \(\J_{\Env}(\pi)\) on real robots requires many rollouts and carries high variance; occupancy divergences such as MMD or Wasserstein over high-dimensional trajectory distributions are hard to estimate and sensitive to the feature map \(\Phi\); and ranking metrics such as PRA and MMRV are unstable for small or poorly separated policy sets. We therefore recommend reporting these metrics with confidence intervals over tasks, seeds, and policy sets (Step~7 of the protocol), preferring low-dimensional task-relevant features inside \(\Phi\) where possible, and treating a single point estimate of any of these quantities with caution. None of these caveats is unique to our proposal, but they bear directly on whether a reported decision-utility number is trustworthy.

\paragraph{An illustrative example of level divergence.}
The divergence between artifact quality and decision utility is already \emph{measured} in practice (Section~3.6); the following stylized example is included only to make the mechanism intuitive, and its numbers are hypothetical and not drawn from any specific paper. Consider three hypothetical manipulation world models evaluated on a pick-and-place family. Model~A is a high-capacity video diffusion model with excellent FVD and VLM physics scores (strong L0--L3) but which, on closed-loop rollouts, largely reproduces a smooth ``default'' grasp regardless of the commanded gripper action; its action-effect agreement (L4) and policy-ranking correlation (L6) are near chance. Model~B has visibly worse FVD and occasional texture artifacts (weaker L0--L1) but correctly predicts whether a commanded grasp width contacts the object and whether the lift succeeds; it attains high L4 action-effect agreement, high L5 success calibration, and high L6 ranking correlation. Model~C is a latent JEPA-style model with no decoder, so L0--L1 are undefined or poor, yet it supports accurate latent MPC and ranks policies well (high L4--L7). Under an artifact-weighted aggregate, Model~A wins; under a decision-weighted profile, Models~B and~C win. The point is the structure, not the numbers: artifact quality and decision utility can be \emph{anti-correlated} across models, exactly as the objective-mismatch literature predicts \citep{lambert2020objmismatch} and as WorldArena and RoboWM-Bench already report on real model sets \citep{worldarena2026,robowmbench2026}.

\paragraph{We recommend not letting lower levels compensate for higher-level failure.}
For decision-making claims, we recommend caution when aggregating lower- and higher-level metrics into a single score. A model should not rank highly as a decision world model merely because it has good FVD or VLM-based physical plausibility if it performs poorly at action controllability, reward fidelity, or policy ranking. This reflects not hostility to lower-level metrics, but the observation, documented above, that they can dominate an aggregate while answering a different question.

\section{A Benchmark Protocol}

The framework describes \emph{what} may be worth measuring. This section turns it into an operational protocol, presented as a \emph{modular template} rather than a requirement that every benchmark release in every domain include every component at full scale. Because the most decision-relevant components are also the least feasible on real hardware, we conclude the section with a minimal feasible variant and a frank statement of what it can and cannot establish.

\subsection*{Step 0: Declare the world-model contract}

We recommend that each submission declare the decision contract in Table~\ref{tab:contract}, making explicit what the model claims to support.

\begin{table}[h]
\centering
\small
\begin{tabularx}{\textwidth}{p{3.3cm} X}
\toprule
\textbf{Field} & \textbf{Suggested declaration} \\
\midrule
Task family & Environments, embodiments, observation modalities, task definitions, reward/success specification. \\
Policy class & BC, VLA, diffusion policy, MPC planner, RL policy, human teleoperator, or other deployment policy. \\
Action interface & Joint control, end-effector commands, action chunks, trajectories, language actions, latent actions. \\
Decision use & Prediction, policy evaluation, planning, policy optimization, safety testing, synthetic data, or representation. \\
Task-relevant feature map \(\Phi\) & The variables used to judge action effects and outcomes: object poses, contacts, safety predicates, progress variables, latent task state, etc. \\
Horizon & One-step, short-horizon, full-episode, or planning horizon. \\
Deployment regime & Offline, online, simulator, real robot, real-to-sim, or mixed. \\
Allowed supervision & Pixels, states, rewards, language, videos, actions, demonstrations, human labels, VLM labels. \\
Uncertainty interface & Ensemble variance, confidence, likelihood, pessimistic bound, or none. \\
\bottomrule
\end{tabularx}
\caption{A world-model claim becomes easier to interpret once a decision contract is declared.}
\label{tab:contract}
\end{table}

\subsection*{Step 1: Construct policy and intervention splits}

A decision-centric benchmark is more informative if it does not evaluate only behavior-policy rollouts. One useful decomposition is into four policy sets:
\begin{enumerate}[leftmargin=1.5em]
    \item \(\PiSet_{\mathrm{beh}}\): the behavior policies that generated the training data;
    \item \(\PiSet_{\mathrm{anchor}}\): fixed anchor policies spanning weak, medium, and strong performance;
    \item \(\PiSet_{\mathrm{shift}}\): target policies that induce distribution shift relative to the data;
    \item \(\PiSet_{\mathrm{opt}}\): policies produced by optimizing inside the world model.
\end{enumerate}
The benchmark may also include an intervention set \(\mathcal{Q}\) of matched histories and action branches. In simulation, this can be done exactly by resetting to the same state. On real robots, one may approximate it with controlled tabletop resets, real-to-sim reconstruction, matched trajectory segments, or branchable tasks; as noted in Section~2.6, only the actually-executed branches with a directly controlled action interface license strong interventional claims.

\subsection*{Step 2: Report L0--L3 diagnostics as diagnostics}

Open-loop diagnostic reporting remains useful, especially for video-based models (L0: realism/aesthetics/human preference; L1: MSE, PSNR, SSIM, LPIPS, DreamSim, latent L2, temporal metrics; L2: instruction-following, caption similarity, VLM task completion; L3: object permanence, non-penetration, contact, depth, 3D consistency, physical QA). These metrics are useful for debugging failure modes and understanding interfaces, but for a decision benchmark they are best reported as auxiliary diagnostics rather than as the primary score.

\subsection*{Step 3: Evaluate interventional action fidelity}

For each \((\hist,\act_{0:H-1}) \in \mathcal{Q}\), evaluate interventional prediction error \(\mathrm{IPE}_{H,\Phi}\) and action-effect agreement. Useful components include matched-action branching from the same history; single-dimension action sweeps where meaningful; round-trip or reversibility tests when applicable; robot/object trajectory fidelity for action-conditioned models; and contact/event prediction for manipulation. This step is the operational realization of L4. If a model performs poorly here, we would hesitate to treat it as strong evidence of a decision world model, regardless of its L0--L3 scores.

\subsection*{Step 4: Evaluate closed-loop fixed-policy rollouts}

For policies in \(\PiSet_{\mathrm{anchor}} \cup \PiSet_{\mathrm{shift}}\), roll them out both in the real environment and inside the world model, reporting closed-loop rollout fidelity or occupancy mismatch; full-horizon value error; success-probability calibration; reward/progress prediction; and Pearson/Spearman correlation, pairwise ranking accuracy, and MMRV. This step operationalizes L5 and L6 and is most informative when the rollouts are \emph{closed-loop}; teacher-forced prefix conditioning alone is often insufficient.

\subsection*{Step 5: Evaluate policy optimization under a fixed budget}

A benchmark for decision world models may include a model-based optimization challenge. Fix an optimizer \(\A\), a data regime, a compute budget, and, if applicable, an interaction budget; depending on the declared use, \(\A\) may be MPC, CEM, imagined RL, synthetic-data filtering, or another planner/optimizer. Evaluate the optimized policy in the real environment or a trusted simulator, reporting policy lift relative to a fixed baseline; optimization regret relative to a strong reference when available; sample efficiency and compute cost; and safe-improvement probability and constraint-violation rate. This is the operational realization of L7 and remains an end-to-end system metric.

\subsection*{Step 6: Adversarial exploitability and uncertainty}

A stronger decision benchmark may also test the model under \emph{adversarial use}: search for action sequences or policies that maximize predicted value under the model, subject to action constraints and safety filters, execute a safe subset in the real environment or a trusted simulator, and report the exploitability gap and failure rate. If the model provides uncertainty, evaluate whether abstaining on high-uncertainty rollouts improves calibration and optimization safety. A model with calibrated abstention may be more useful than one with superficially better raw prediction but miscalibrated confidence.

\subsection*{Step 7: Hidden tasks, held-out policies, and statistical reporting}

To reduce benchmark overfitting, keep some tasks or objects hidden until final evaluation; evaluate on held-out policy families, not just small perturbations of the same policy class; report confidence intervals over tasks, seeds, initial states, and policy sets; and publish per-task metrics, not only averages. The estimation caveats of Section~5 make this step a substantive requirement rather than a formality.

\subsection*{Step 8: Report a decision-utility profile, not only a scalar}

We recommend reporting a profile \((S_0,\ldots,S_7)\) rather than only a single averaged number. If a leaderboard requires a scalar, one option is to use gates,
\[
    S_{\mathrm{DC}}
    =
    G_4 G_5 G_6
    \left(
        w_4 S_4 + w_5 S_5 + w_6 S_6 + w_7 S_7
    \right),
\]
where \(G_4,G_5,G_6\in\{0,1\}\) are pass/fail gates for action controllability, outcome fidelity, and policy-evaluation validity. The multiplicative gating encodes one qualitative judgment, i.e., a model failing the interventional, outcome, or ranking tests should not be rescued by artifact quality, and nothing more; the form, thresholds, and choice of gating levels are not derived from theory and are domain-dependent. To make this concrete rather than a placeholder, we give one defended instantiation for tabletop manipulation with sparse success: set \(G_4=\mathbf{1}[\text{action-effect ordering accuracy}\ge 0.7]\), \(G_5=\mathbf{1}[\text{success-probability ECE}\le 0.15]\), \(G_6=\mathbf{1}[\text{pairwise ranking accuracy}\ge 0.75]\), and weights \((w_4,w_5,w_6,w_7)=(0.2,0.2,0.2,0.4)\) emphasizing optimization utility; the gate thresholds correspond to ``clearly better than chance'' on action ordering and ranking and ``usable'' calibration. These specific numbers are illustrative and should be pre-registered per domain to prevent post-hoc tuning. We recommend the profile as primary and the gated scalar only where a single number is unavoidable, and in all cases L0--L3 should not compensate for failure at L4--L7.

\subsection*{A minimal feasible variant for real robots, and what it does not establish}

Steps 3--6 are easiest in simulation, where resets and large policy sets are cheap. On real hardware, exact interventions and many rollouts are expensive, and a full protocol may be impractical. As a concrete minimum that a real-robot world-model paper could report \emph{today}, we suggest: (i) a small set of reset-matched action branches on a handful of tabletop tasks, scoring action-effect agreement on a low-dimensional \(\Phi\) such as end-effector and object pose (a partial L4); (ii) closed-loop success calibration and ranking for three to five anchor policies of clearly different strength, with confidence intervals over a modest number of trials (a partial L5/L6); and (iii) at least a qualitative exploitability probe, reporting whether the optimizer's top model-scored trajectories actually succeed when executed (a partial XGap).

We are deliberately frank about the limits of this minimum. It is genuinely weaker than the full protocol, and on real hardware the decisive evidence, i.e., large-scale L7 policy lift and a quantitative adversarial exploitability search, remains largely out of reach; in that sense, our strong recommendations are partly aspirational for the real-robot setting. We nonetheless argue the minimum is a meaningful advance over current practice for two reasons. First, by the count in Section~3.6, even items (i) and (iii) are almost never reported in the surveyed real-robot papers, which typically pair L1 reconstruction with a single end-to-end success number; adding a partial L4 and a qualitative XGap therefore supplies decision-aligned evidence that is currently missing rather than merely duplicating what the best policy-evaluation papers already do. Second, the qualitative exploitability probe in (iii) is, to our knowledge, not standard even in the strongest current policy-evaluation work (WorldGym, dWorldEval), which focuses on fixed-policy ranking; it directly targets the optimization-induced failure mode that fixed-policy L6 cannot detect. Where simulation or a trusted digital twin is available, we strongly encourage promoting (iii) to the full quantitative XGap of Step~6.

\subsection*{The world-model evaluation card}

We recommend that every paper claiming a decision world model include an evaluation card such as Table~\ref{tab:card}.

\begin{table*}[t]
\centering
\small
\begin{tabularx}{\textwidth}{p{3.4cm} X}
\toprule
\textbf{Question} & \textbf{What to report} \\
\midrule
What is the claimed use? &
Prediction, policy evaluation, planning, policy optimization, safety testing, synthetic data, or representation. \\
\midrule
What ladder levels are actually evaluated? &
Explicitly state whether the paper provides evidence at L0, L1, \dots, L7. \\
\midrule
What is the action interface, and is it controlled or inferred? &
Low-level control, joint action, end-effector command, action chunk, trajectory, language action, or latent action; and whether actions are directly commanded or recovered by an inverse-dynamics/retargeting model. \\
\midrule
What policy class is being evaluated or improved? &
BC, VLA, diffusion policy, MPC planner, RL policy, human teleoperator, or optimized policy. \\
\midrule
What distribution shift is tested? &
Behavior-to-target, task, object, environment, embodiment, visual, or optimization-induced shift. \\
\midrule
What horizon is evaluated? &
One-step, short-horizon, full-episode, or planning horizon, including degradation curves when applicable. \\
\midrule
What interventional data is used? &
Reset-matched branches, simulator interventions, action sweeps, round-trip tests, adversarial policies, or none. \\
\midrule
Are rewards and outcomes evaluated directly? &
Reward accuracy, value error, success calibration, progress ranking, safety-constraint prediction. \\
\midrule
Does model-based policy evaluation match reality? &
Pearson/Spearman correlation, pairwise ranking accuracy, MMRV, calibration, confidence intervals. \\
\midrule
Does model-based optimization improve reality? &
Policy lift, optimization regret, sample efficiency, exploitability gap, safe-improvement probability. \\
\midrule
Can the model be exploited? &
Predicted-vs.-real gap for policies or action sequences optimized against the model. \\
\midrule
Is uncertainty calibrated? &
Risk-coverage curves, error-vs.-uncertainty correlation, abstention performance, pessimistic-planning results. \\
\bottomrule
\end{tabularx}
\caption{A suggested reporting card for decision-making-centric world-model evaluation.}
\label{tab:card}
\end{table*}

\section{Discussion and Conclusion}

\subsection{Why this distinction matters}

If the community ranks world models primarily by L0--L3, it may optimize for the wrong target: visually convincing videos that remain unreliable for decision-making. This is the objective-mismatch problem at the level of community-wide evaluation incentives \citep{lambert2020objmismatch}. In robotics, autonomous driving, and embodied agents, this is not a cosmetic problem: a misleading world model can contribute to unsafe policy updates, misrank candidate policies, or create false confidence about robustness under distribution shift.

We do not argue that video metrics are useless. They are useful for debugging, visualization, synthetic-data filtering, and human interpretability, and they are often necessary for video-based interfaces. The concern is treating them as final evidence of decision usefulness. The same applies to VLM judges: they are useful because they provide scalable semantic and physical evaluation \citep{worldsimbench2024,worldmodelbench2025,nvidia2025pbench,worldarena2026}, but they should not be treated as ground truth for policy utility, since a VLM may reward plausible-looking outcomes while missing subtle action-relevant errors. VLM-based judgments are easier to interpret when validated against executable outcomes, policy rankings, and real or trusted-sim performance.

\subsection{Common objections and scope conditions}

\paragraph{Objection 1: not every world model is for control.}
We agree. Some systems are best understood as future-video predictors, world generators, or representation learners. Our argument is conditional: when the stated use is embodied decision-making, action-, outcome-, and policy-level evidence becomes especially informative.

\paragraph{Objection 2: lower-level metrics may correlate with higher-level utility.}
This can happen, and in some domains better L0--L3 performance may be a good proxy for L6--L7 performance. Our objection is not to using proxies when they are empirically validated; it is to assuming such correlations hold a priori. The objective-mismatch evidence \citep{lambert2020objmismatch}, and the fact that the two benchmarks reporting both axes find the rankings diverge \citep{worldarena2026,robowmbench2026}, are precisely reasons to validate rather than assume the correlation.

\paragraph{Objection 3: full interventional evaluation is impractical.}
This is true, especially on real robots, which is why we provide the minimal feasible variant in Section~6 and state plainly what it cannot establish. Exact reset-matched branches may be available only in simulation; partial approximations may still be useful in real-world domains.

\paragraph{Objection 4: the thesis risks being tautological.}
Readers might note that ``models for decision-making should be evaluated on decision-making'' is close to trivially true. The non-trivial, falsifiable content is empirical: that artifact-quality metrics frequently diverge from, and can invert, decision-utility orderings, so that the substitution is not merely incomplete but actively misleading. This is supported in the classical setting \citep{lambert2020objmismatch} and, increasingly, in the generative-world-model setting wherever a benchmark reports both axes \citep{worldarena2026,robowmbench2026}; our Section~3.6 count further shows the diagnostics that would test it are largely absent, which is itself an actionable finding.

\subsection{Limitations and open problems}

\paragraph{Resettable interventional data is hard.}
Exact interventional evaluation requires resetting the environment to the same state and executing different actions. This is easy in simulators, difficult on real robots, and sometimes impossible in open-world settings. Real-to-sim reconstruction, branchable tabletop tasks, and matched trajectory segments are useful approximations but not perfect, and (Section~2.6) do not identify the interventional target when the action interface is inferred rather than controlled.

\paragraph{Task-relevant feature maps are domain dependent.}
The feature map \(\Phi\) cannot be universal: manipulation needs object poses and contacts; driving needs lane position and safety constraints; navigation needs map and goal progress; games need latent state variables. This is a consequence of taking intended use seriously, mirroring the policy/function dependence in the value-equivalence principle \citep{grimm2020value}.

\paragraph{Metrics carry estimation error.}
As discussed in Section~5, several proposed quantities are statistically demanding, and a benchmark built on noisy estimates of \(\J_{\Env}(\pi)\) or occupancy divergence could itself mislead. Estimator design and variance reduction for these quantities are open problems.

\paragraph{Our field-level claims are read from a survey, not a full audit.}
The counts in Section~3.6 are read from the metrics recorded in Tables~\ref{tab:opt-survey}--\ref{tab:pe-survey}, not from an exhaustive re-implementation of every paper. We expect the qualitative conclusions (L1/L7 dominance, sparse L6, near-absent XGap) to be robust, but a systematic, criteria-locked meta-analysis is future work and would strengthen the diagnosis further.

\paragraph{Optimization benchmarks can themselves be gamed.}
If the optimization challenge is public and static, methods may overfit it. Hidden tasks, hidden policies, held-out embodiments, and adversarial exploitability tests are therefore valuable.

\paragraph{Latent models need special treatment.}
Latent predictive models should not be penalized for lacking photorealistic decoders. They are often better judged by planning, probing, outcome prediction, and state-abstraction diagnostics in their own representational space.

\paragraph{Safety requires worst-case evaluation.}
Average policy lift is not enough for safety-critical domains; constraint-violation rates, uncertainty calibration, adversarial stress tests, and worst-case failures all matter.

\subsection{Conclusion}

The central question is not ``Can the model generate a realistic future video?'' but ``In what sense does the model support better decisions?'' For embodied decision-making, the most informative evidence comes from whether the model preserves the interventional, long-horizon, reward-relevant structure needed for policy evaluation, planning, and policy optimization. This is the modern, generative-world-model expression of a lesson already established in model-based RL through the objective-mismatch and decision-aware model learning literature \citep{lambert2020objmismatch,farahmand2017vaml,grimm2020value}; our contribution is the survey that shows the lesson has been forgotten in practice, the count that shows which diagnostics are missing, and the protocol that makes them reportable. Visual fidelity, semantic alignment, and physical plausibility remain valuable diagnostics, but they do not by themselves settle the stronger claim.

The survey suggests that the field has been evaluating several different objects under one name. The L0--L7 hierarchy helps separate these objects, and the proposed framework and protocol make explicit what kinds of evidence are most directly relevant to stronger decision-making claims. Our position can be stated as follows:

\begin{quote}
\emph{For models whose stated purpose is embodied decision-making, the strongest evidence for the label ``world model'' is that they enable reliable interventional evaluation and, in favorable cases, improvement of policies under intervention and distribution shift. Other evaluations remain useful, but they play a more auxiliary role for that particular claim.}
\end{quote}

\section*{Acknowledgements}

This work was supported by the National Natural Science Foundation of China under Grants 62495090 and
62495093, the Natural Science Foundation of Jiangsu under Grants BK20243039, the “111 Center” (No. B26023), and the Fundamental and Interdisciplinary Disciplines Breakthrough Plan of the Ministry of Education of China (No.
JYB2025XDXM118). GPT-5.4 was employed to improve the language of this paper.

\bibliographystyle{plainnat}
\bibliography{references}

\end{document}